\UseRawInputEncoding
\documentclass[10pt,twocolumn,letterpaper]{article} 

\usepackage{avss}
\usepackage{times}
\usepackage{epsfig}
\usepackage{subcaption}
\usepackage{graphicx}
\usepackage{amsmath}
\usepackage{amssymb}
\usepackage{tabularx}
\usepackage{xcolor}
\usepackage{bbm}
\usepackage{algorithm}
\usepackage{algpseudocode}
\newif\ifcomments
\commentstrue
\commentsfalse

\ifcomments
    \usepackage{ulem}
    \def\mc#1{{\color{blue} [\textbf{MC:} #1]}}
    
    \def\jlk#1{{\color{red} [\textbf{JLK:} #1]}}
    
    \def\picomment#1{{$\!$\color{magenta} [PI: #1]}}

\else 
    \def\mc#1{}
    
    \def\jlk#1{}
    
    \def\picomment#1{}
    
    \def\jbedit#1{}
    
\fi

\usepackage[pagebackref=true,breaklinks=true,letterpaper=true,colorlinks,bookmarks=false]{hyperref}

\avssfinalcopy 

\def\avssPaperID{35} 

\ifavssfinal\fi
\begin{document}

\title{Spatio-Visual Fusion-Based Person Re-Identification \\
for Overhead Fisheye Images}

\author{Mertcan Cokbas, Prakash Ishwar, Janusz Konrad\thanks{This work was supported by ARPA-E (agreement DE-AR0000944)
}\\
Department of Electrical and Computer Engineering, Boston University\\
8 Saint Mary's Street, Boston, MA 02215\\
{\tt\small [mcokbas, pi, jkonrad]@bu.edu}
}

\maketitle

\begin{abstract}
Person re-identification (PRID) has been thoroughly researched in typical surveillance scenarios where various scenes are monitored by side-mounted, rectilinear-lens cameras.
To date, few methods have been proposed for fisheye cameras mounted overhead and their performance is lacking. In order to close this performance gap, 
we propose a multi-feature framework for fisheye PRID where we combine deep-learning, color-based and location-based features by means of novel feature fusion. We evaluate the performance of our framework for various feature combinations on FRIDA, a public fisheye PRID dataset. The results demonstrate that our multi-feature approach outperforms recent appearance-based deep-learning methods by almost 18\% points and location-based methods by almost 3\% points in matching accuracy. We also demonstrate the potential application of the proposed PRID framework to people counting in large, crowded indoor spaces.
\end{abstract}

\section{Introduction}
\label{sec:introduction}
Detecting, tracking and counting people indoors is an enabling technology for energy-waste reduction (e.g., HVAC, lighting), space management (e.g., quantifying underutilization) and in emergency situations (e.g., fire/chemical hazard, active shooter).
While a number of approaches have been developed to detect and track people indoors, {\it overhead fisheye} cameras have recently emerged as a compelling alternative. With their vast field of view ($360^\circ$ horizontally and $180^\circ$ vertically), fewer fisheye cameras are needed to cover a large space than commonly-used standard rectilinear-lens cameras, thus reducing system deployment costs.

In settings involving multiple cameras, the task of accurately re-identifying the same person (person re-identification or PRID) in the frames captured by different cameras is important for certain applications such as people counting or tracking.
While the PRID problem has been thoroughly researched for typical rectilinear cameras~\cite{PRID_overview}, little work exits on PRID using overhead fisheye cameras which introduce unique challenges due to the overhead viewpoint and unusual lens geometry.
To the best of our knowledge, only 4 works have been published to date focusing on fisheye PRID: \cite{fisheye_PRID_Barman,fisheye_PRID_Blott,josh,FRIDA} each with its own limitations (from viewpoint restrictions to the need for camera calibration). Furthermore, each approach exploits only a single trait, either a hand-crafted or deep-learning feature, or a location constraint.
\begin{figure*}[htb]
\centerline{\includegraphics[width=0.38\textwidth]
{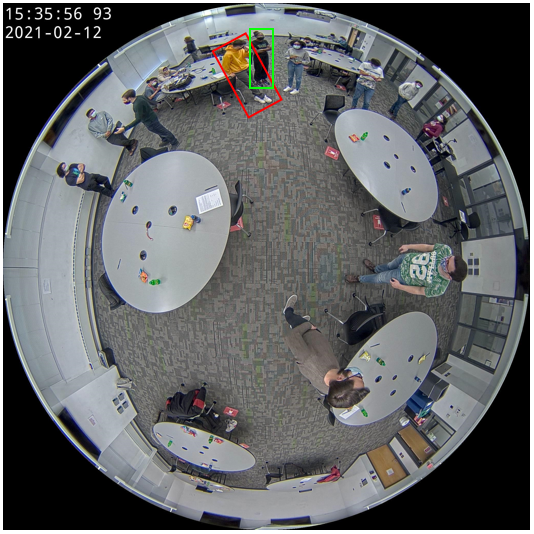}\quad
\includegraphics[width=0.38\textwidth]{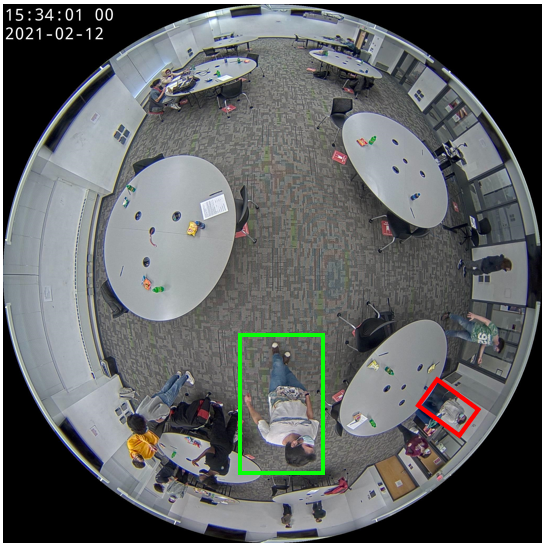}}
\medskip

\centerline{\includegraphics[width=0.38\textwidth]{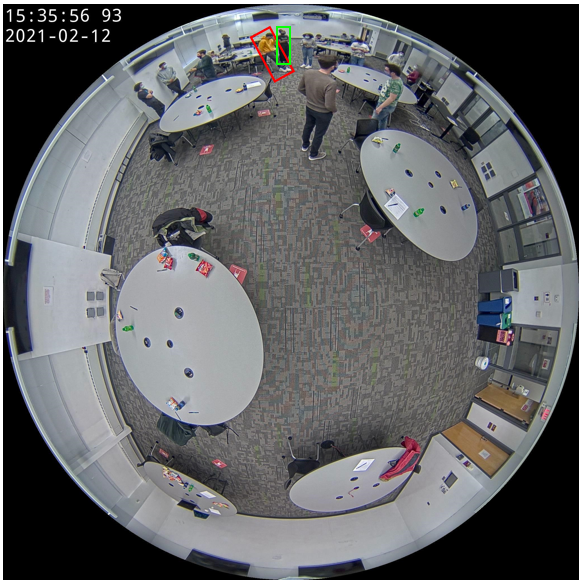}\quad
\includegraphics[width=0.38\textwidth]{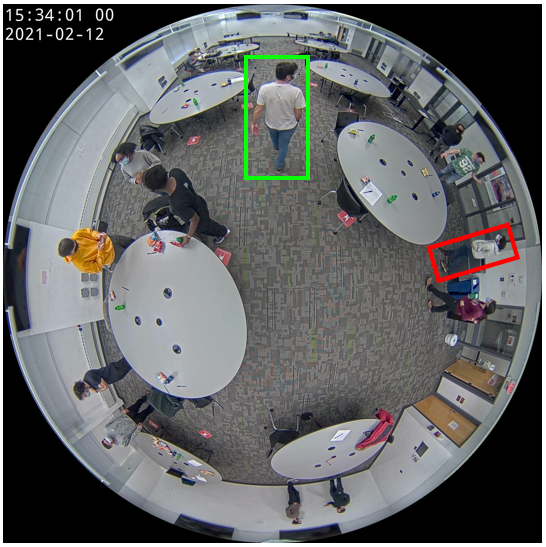}}
\centerline{Location ambiguity\hskip 0.26\textwidth Appearance ambiguity}

\vglue 0.1cm
\caption{Illustration of location and appearance ambiguity. When people are very close to each other, distinct color of clothing and/or body shape/features may help resolve location ambiguity (left column). When people have very similar appearance (e.g., light-gray T-shirts and dark pants), knowing their location may help resolve appearance ambiguity (right column).} \label{fig:motivation}
\end{figure*}

In this paper, we propose a framework for PRID from time-synchronized overhead fisheye cameras with overlapping fields of view (FOVs) that combines appearance and location information to improve performance.
When people are very close to each other, matching identities using location information is often perilous due to location-estimation errors \cite{josh}. In this case, visual characteristics of a person may help disambiguate any confusion (see an example in the left column of Figure \ref{fig:motivation}). On the other hand, when people look very similar (color of clothing, body shape), appearance-based methods frequently fail to differentiate between them. However, 3-D location in a room can be occupied by one person only, so location information should help with correct matching (see an example in the right column of Figure \ref{fig:motivation}). Only when two similarly looking people are next to each other this approach would fail, but then a human eye would likely fail as well.

For appearance information, we propose to combine hand-crafted and learning-based features. As the former, we use a color histogram computed from bounding box around each person.
As the latter, we use an embedding from state-of-the-art PRID deep-learning method developed for rectilinear cameras and fine-tuned on fisheye data (domain transfer). In order to obtain person-location information, we use the method developed in \cite{josh}. We use Na\"{i}ve Bayes fusion to combine the color, embedding and location data by converting each to a similarity score (between two identities) and multiplying the scores after suitable normalization. We perform identity matching by a greedy algorithm. 

Similarly to traditional PRID research, we use human-annotated ground-truth bounding boxes for each person to evaluate different PRID methods.
%
Evaluation on a recent fisheye PRID dataset \cite{FRIDA} demonstrates that location-based features deliver accuracy slightly over 94\%. Appearance-based features, however, manage to achieve only 64-80\% in matching accuracy. Despite their lower performance, when they are combined with location-based features the accuracy exceeds 97\%. 

%
We conclude our investigation by exploring the use of overhead fisheye PRID for people counting in a realistic scenario where room occupancy is high and bounding boxes for people are produced by a state-of-the-art person-detection algorithm for overhead fisheye images (instead of idealized human-annotated ground-truth boxes). Our experimental results demonstrate that without PRID the people counts are either significantly overestimated or significantly underestimated.

To summarize, our main contributions are:
\begin{itemize}\itemsep 0em
    \item A framework for multi-feature person re-iden\-tification using time-synchronized overhead fisheye cameras with overlapping FOVs. 
    \item A novel probabilistic feature-fusion method for identity matching. 
    \item An ablation study of the importance of individual features and of various combinations thereof on a recent fisheye PRID dataset \cite{FRIDA}.
    \item Demonstration of the potential application of multi-feature PRID for real-life people counting in a high-occupancy scenario.
\end{itemize}

\section{RELATED WORK}
\label{sec:related_work}

Traditional PRID methods developed to date have considered matching identities captured at {\it different time instants} (asynchronously) by multiple cameras with {\it no FOV overlap}. In this case, a query ID from one camera is allowed to have multiple matches in the gallery set that is extracted from other cameras. In principle, asynchronous PRID from cameras {\it with overlapping FOVs} would fall into the traditional PRID category but we are not aware of any such work.

An alternative is a different type of PRID, that we call {\it cross-frame PRID}, where the ID matching is performed between video frames captured at the {\it same time instant} (synchronously) by two cameras with {\it overlapping FOVs}. In this case, person-images from one camera's video frame are considered query elements while those from the other camera's synchronous video frame are considered to be the gallery set. Consequently, a query ID is allowed to have only {\it one correct match} (or none, in case of occlusion) in the gallery set. Note, that synchronous PRID from cameras with no FOV overlap is not possible since a person would appear only in a single camera view.

We summarize these differences in  Table~\ref{tab:Problem_Setup_Difference} and review both traditional and cross-frame PRID for rectilinear and fisheye cameras below.

\begin{table*}[!hbt]
    \centering
    \begin{tabular}{|c|c|c|c|}\hline
  \textbf{PRID setting}  & \textbf{FOV} & \textbf{Re-ID timing} & \textbf{Gallery samples per identity} \\ \hline
%
    Traditional & Non-Overlapping & Asynchronous & Multiple \\ 
     {\bf Cross-Frame} (this paper) & Overlapping & Synchronous & Single/None \\   \hline  
    \end{tabular}
    \caption{Key differences between traditional and cross-frame PRID. 
    }
    \label{tab:Problem_Setup_Difference}
\end{table*}

\subsection{Traditional Rectilinear PRID} 
Traditional PRID methods for side-mounted rectilinear-lens cameras have been thoroughly researched to date.
Early methods were model-based and relied on hand-crafted features, for example Gaussian descriptors \cite{gaussian_descriptor},
or color information  \cite{saliency_color,LF_symmetry}.
A detailed review of such approaches can be found in \cite{trad_PRID_review}.

As the applications of PRID grew, large datasets were introduced. Methods that rely on hand-crafted features were found to be not robust or failed to generalize well to these new datasets. At the same time, these datasets enabled researchers to develop sophisticated deep-learning methods that eventually exhibited superior performance compared to classical PRID methods \cite{PRID_overview}. 
 This trend is similar to one observed in other computer-vision tasks such as object detection, recognition, and tracking where deep-learning methods tend to outperform model-based approaches.
Today, the best-performing PRID methods indeed rely on deep learning \cite{ABD,CTL,VA_ReID,Pyramid,PCB}. Recently, several of these methods have been evaluated on a fisheye PRID dataset \cite{FRIDA} with the conclusion that their performance drops significantly when applied directly to overhead fisheye images. Among these algorithms CTL, based on a triplet loss \cite{CTL}, outperforms other methods.

Among the datasets developed for traditional PRID perhaps the most frequently-used are CUHK03 \cite{CUHK03} and Market-1501 \cite{Market-1501}, both recorded by side-mounted rectilinear-lens cameras with no FOV overlap.

\subsection{Traditional Fisheye PRID}
Recently, traditional PRID methods have been extended to overhead fisheye cameras.
%
%
The early work in \cite{fisheye_PRID_Barman} considers only matching people who are at a similar distance from each camera, thus assuring similar body size (and, potentially, similar viewpoint). 
Another work \cite{fisheye_PRID_Blott} uses tracking to extract 3 distinct viewpoints (back, side, front) that are subsequently jointly matched between cameras. This requires reliable tracking and visibility of each person from 3 distinct viewpoints -- neither can be guaranteed. Such distance or tracking constraints severely limit the applicability of these methods.
Moreover, both works report results on non-public fisheye data only. 

We note that neither method is suitable for cross-frame PRID. The assumption of similar distance from different cameras enforced in \cite{fisheye_PRID_Barman} would allow cross-frame PRID only in a very small range of locations in the monitored space. Also, the need for 3 distinct views from each camera, required in \cite{fisheye_PRID_Blott}, is not applicable in cross-frame PRID since it is performed at single time instant thus allowing only one view.

\subsection{Cross-Frame Rectilinear PRID} \label{sec:cross-frame-rectilinear}

Cross-frame PRID finds fewer applications than traditional PRID since it
requires multiple time-synchronized cameras to monitor the same space (increased cost and complexity).
It has been primarily used in people detection and tracking.

Fleuret \textit{et al.}\ \cite{Fleuret_2014} applied cross-frame PRID to improve tracking in sporting events. Their approach is appearance-based and uses jersey color, jersey number and facial features for re-identification between concurrent rectilinear camera views. Hu \textit{et al.}~ \cite{surgical_room} also focused on improving tracking of people but in a smaller space of a hospital operating room. They first perform 3-D tracking of each person's skeleton from calibrated rectilinear cameras and then re-identify people between different camera views by clustering 3D trajectories. Each 3D-trajectory cluster is processed to produce a robust 3D trajectory for each person. This approach uses no appearance features for re-identification. Finally, Wang {\it et al.} \cite{Wang2021} proposed precise localization of people indoors using up to 8 calibrated, time-synchronized, rectilinear cameras. They estimated 2-D skeletons of people using OpenPose library and projected them to 3-D space for distance-based clustering. This 3-D skeleton clustering is a form of location-based cross-frame PRID.


In all these studies, cameras have a side view of the scene which can cause occlusions and no-match scenarios. To address this, one possible solution is to mount cameras overhead and point them down. However, due to a relatively-narrow FOV of rectilinear cameras, this solution would be impractical as it would require many cameras to cover a large space. 

\subsection{Cross-Frame Fisheye PRID}


A more practical approach to reducing occlusions and increasing area coverage is to use overhead fisheye cameras for their wide FOV. However, appearance matching 
is more challenging in this case on account of fisheye-lens geometric distortions.

The first cross-frame fisheye PRID method was developed in \cite{josh}. Instead of matching the appearance of people, they performed matching of people's locations between camera views. In this case, it is critical that cameras be precisely time-synchronized. Evaluated on a non-public dataset, the proposed method performed very well but required careful camera calibration.

While the work in \cite{FRIDA} did not introduce any new fisheye PRID method, it introduced a new PRID dataset captured by time-synchronized overhead fisheye cameras with overlapping FOVs and made it publicly available. The paper also evaluated several state-of-the-art traditional rectilinear PRID algorithms adapted to fisheye PRID on this dataset.

We note that the methods of cross-frame rectilinear PRID discussed in Section~\ref{sec:cross-frame-rectilinear} 
are somewhat specialized to their respective use cases (e.g., recognizing jersey numbers) or cannot be easily generalized to the overhead fisheye camera setting due to the unique lens distortion and viewpoint, e.g., skeletons and facial features cannot be recognized right below the camera.
In this paper, we study cross-frame PRID for fisheye cameras and demonstrate the effectiveness of proposed multi-feature framework on the dataset from \cite{FRIDA}.

\section{METHODOLOGY}

\begin{figure*}[!hbt]
    \centering
    \includegraphics[width=1\linewidth]{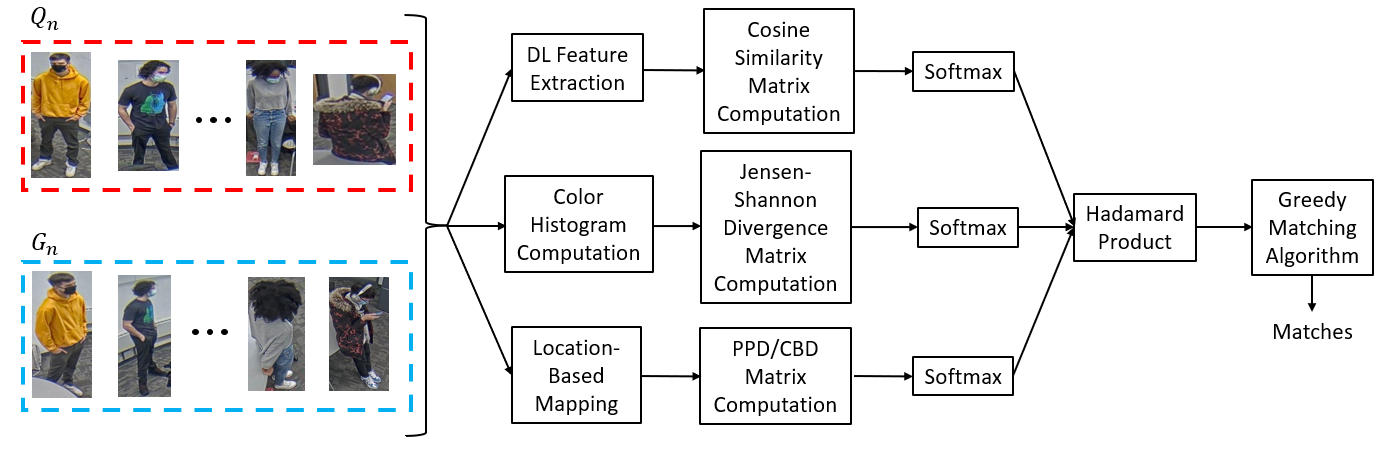}
    \caption{Block diagram of the proposed method.  In the first step, features are extracted from the contents of bounding boxes (for appearance-based methods) or their positions within the frames (for location-based method) for the query and gallery identity sets (which are denoted by $Q_n$ and $G_n$, respectively, for frame number $n$). In the next step, features of query-gallery pairs are converted into either pairwise-similarity or pairwise-dissimilarity scores. They are then normalized via the softmax operation to obtain match-probability matrices for each feature group which are then fused together via the Hadamard product into a fused match-probability matrix. Finally, a greedy sequential matching algorithm based on the fused match-probability matrix is used to produce the query-gallery matches.}
    \label{fig:block_diagram}
\end{figure*}

In order to perform cross-frame PRID between two fisheye frames, we designate one of the frames as a {\it query} frame and the other frame as a {\it gallery} frame. We denote the sets of identities of people in the query and gallery frames at time $t_n$ as $Q_n$ and $G_n$, respectively. We compute a $|Q_n|\times |G_n|$ score matrix where, $|\cdot|$ is the cardinality operator. The score in the $i^{th}$ row and $j^{th}$ column of this matrix represents the similarity between the $i^{th}$ query element and the $j^{th}$ gallery element.
Each score is a combination of 3 scores, each derived from a different type of feature: (1) neural-network embedding; (2) color histogram; and (3) location, as shown in Figure \ref{fig:block_diagram}.
%
%
Below, we discuss how each feature is computed, how it is converted to a pairwise-similarity or -dissimilarity score, how these scores from different features are converted into match-probabilities and  fused together and, finally, how the matching between query and gallery elements is performed based on the fused match-probabilities.

\subsection{Deep-Learning Features and Pairwise-Similarity Scores}
\label{ssec:dlfeatures}

Deep-learning methods perform exceedingly well in many visual inference tasks, including rectilinear-camera PRID. It is only natural to consider features extracted by such methods for fisheye PRID as well.

{\bf Features:} To extract deep-learning features, we pass the content of each bounding box (assumed given) through the CTL deep neural network \cite{CTL}. We opt for CTL since it is the top-performing model among 6 state-of-the-art PRID models evaluated on fisheye data \cite{FRIDA}. We use the output of the last convolutional layer in the CTL model as the feature vector.
%

{\bf Pairwise-Similarity Scores:} We compute cosine similarity between the features of the query and gallery elements to obtain a $|Q_n| \times |G_n|$ pairwise-similarity matrix at time $t_n$. Since cosine similarity is symmetric, its values are unaffected by whether a given camera view is designated as query or gallery. 

\subsection{Color Histograms and Pairwise-Dissimilarity Scores}

\begin{figure}[h]
\centering
    \begin{subfigure}[h]{0.49\textwidth}  
    \centering
            \includegraphics[width=0.76\textwidth]{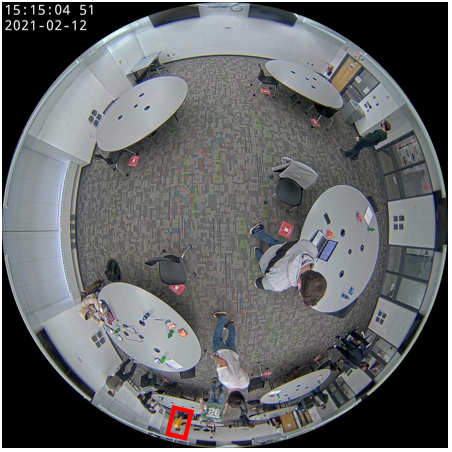}
    \end{subfigure}
    \medskip
    
    \begin{subfigure}[h]{0.49\textwidth}
            \centering
            \includegraphics[width=0.76\textwidth]{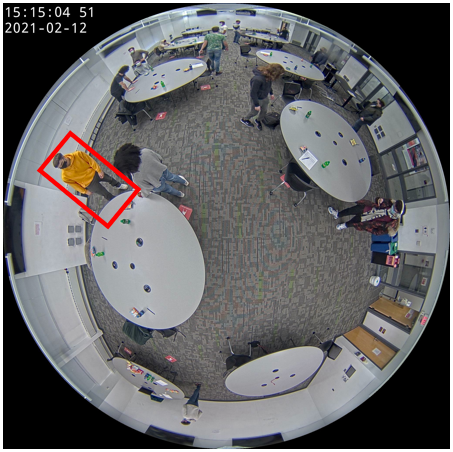}
    \end{subfigure}
    \caption{Illustration of the importance of color information in fisheye PRID. The person in the red bounding box in the top frame is severely shrunk, however the distinctive color of the sweatshirt still allows to distinguish this person from others.}\label{fig:motivation_for_CH}
\end{figure}

In the last decade, deep-learning methods have handily outperformed methods using hand-crafted features in many computer vision tasks and PRID is no exception. However, there exists a performance gap between rectilinear-camera PRID and fisheye-camera PRID. Hand-crafted features that capture prior knowledge could prove useful in closing this peformance gap. The gap is primarily caused by fisheye-lens distortions (and the related huge resolution difference between image center and periphery), and also by dramatic body-viewpoint variations due to the overhead camera placement. One feature that is not impacted by lens distortions and viewpoint is color. No matter how far a person is from the camera, their outfit or skin color does not change (see an example in Figure~\ref{fig:motivation_for_CH}). We leverage this observation, by using color histogram as a hand-crafted identifier of each person.


{\bf Features:} Rather than using a 3-D RGB histogram, we convert RGB values to HSV space and then compute a 1-D histogram of hue H and normalize it to obtain a probability distribution over discretized hue values. In this way, we reduce the impact of illumination variations (change in value V) and color variations under different lighting (change in saturation S).

{\bf Pairwise-Dissimilarity Scores:} If ${\mathbf q}$ and ${\mathbf g}$ are query and gallery hue probability distributions, respectively,
we measure their dissimilarity by calculating the Jensen-Shannon (JS) divergence $D_{JS}$ between them as follows:
\begin{eqnarray*}
D_{JS}({\mathbf q} || {\mathbf g}) &=& \frac{1}{2}D_{KL}({\mathbf q} || {\mathbf p}) + \frac{1}{2}D_{KL}({\mathbf g} || {\mathbf p})
\end{eqnarray*}
where ${\mathbf p} = ({\mathbf q}+{\mathbf g})/2$ and 
$D_{KL}(\cdot||\cdot)$ is the Kullback-Leibler (KL) \cite{KL} divergence between two probability distributions. 

Although both KL divergence and JS divergence are commonly used as measures of distance between two density/distribution functions, we opt for JS divergence because it is symmetric (unlike KL divergence) and it is always finite (the KL divergence $D_{KL}({\mathbf a} || {\mathbf b}) $ is infinity if there is a component $i$ where $b_i = 0$ but $a_i > 0$).
%
Note that the final result is a $|Q_n|\times|G_n|$ divergence matrix (larger values denote larger dissimilarity) unlike in the case of pairwise-similarity matrix in Section~\ref{ssec:dlfeatures}.

\subsection{Location-Based Features and Pairwise-Dissimilarity Scores}
\label{sec:location}

The deep-learning features and color histograms capture visual appearance of people. However, as mentioned earlier, due to the overhead viewpoint and fisheye-lens distortions the appearance of people may change dramatically between cameras. This difficulty was observed and addressed in \cite{josh} for the case of overlapping FOVs. They proposed to use the location of people rather than their appearance for re-identification.
Their method first maps a person's location (center of the bounding box) from one camera's view to another camera's view, and then 
performs matching of locations based on distances.
There are two caveats to this approach. First, both images must be captured at exactly the same time by cameras with overlapping FOVs, which is the case in our problem statement.
Secondly, the mapping of location between camera views requires the knowledge of a person's height.
The issue of a person's height was addressed in \cite{josh} by developing 
various distance measures between locations, either based on an average person's height or a range of typical heights.

{\bf Features:} We use two distances proposed in \cite{josh}, specifically the Point-to-Point Distance (PPD), which is computationally efficient, and the Count-Based Distance (CBD), which is best-performing but significantly more complex.

{\bf Pairwise-Similarity Scores:} 
Since neither PPD nor CBD is symmetric, we compute a symmetric pairwise-dissimilarity matrix from them as follows.
We swap the query and gallery designations of cameras, and compute another distance matrix. We transpose this matrix and add it to the first one to arrive at the final pairwise-dissimilarity matrix. For more details please see \cite{josh}.

\subsection{Fusion of Features}
\label{sec:fusion_of_features}

While each feature type can be used individually to perform PRID, our motivation is to leverage appearance features (deep-learning and color histogram) to disambiguate location uncertainty while relying on location to differentiate individuals with similar appearance. Ideally, one would select suitable features for each potential match, but this is a more difficult problem that we will consider in future research.

In this work, we combine features for all potential matches in the same way by means of a probabilistic information fusion mechanism. Recall that for each feature type and query/gallery sets we obtain a $|Q_n|\times |G_n|$ pairwise-similarity or pairwise-dissimilarity matrix. We normalize each row of either matrix by applying the {\it softmax} operator with positive sign in the exponent to pairwise-similarity matrices and with negative sign to pairwise-dissimilarity matrices. 
This converts both types of matrices into row-stochastic matrices where each row represents the conditional probabilities of gallery elements (columns) matching a given query (row). 

Finally, we use the Na\"{i}ve Bayes methodology to fuse conditional probabilities from different features by taking the Hadamard product (element-wise multiplication) of the conditional probability matrices of different features to obtain the final match-probability matrix (see Figure~\ref{fig:block_diagram}).

\subsection{ 
Matching Algorithm 
}
\label{ssec:greedy}

Regardless of whether each feature type is used separately or is combined with one or two other feature types, the final match-probability matrix contains elements that describe the normalized degree of similarity between a query identity and a gallery identity. In order to match query and gallery identities, one could maximize the sum of logarithms of match-probabilities (or equivalently the product of match-probabilities) for all possible matches via the Hungarian algorithm, but this is computationally expensive. Another option is to perform greedy matching in the following steps: 
\begin{enumerate}
\item pick the maximum match-probability in the matrix and assume the corresponding query and gallery identities match,
\item remove the row and column of the matching identities from the matrix,
\item repeat the first 2 steps until no more matches are possible. 
\end{enumerate}



This procedure yields a \textit{matching} that associates query identities to  gallery identities. A detailed pseudo-code for a possible implementation of the greedy algorithm is provided in Algorithm~\ref{alg:greedy}. In this code instead of deleting rows and columns of matched pairs, we assign a value of $-\infty$ to them. This is a convenient method to keep track of identities without changing the size of the match-probability matrix.
The \textit{probability of the matching} is taken to be the product of the match-probabilities for all the query-gallery matches in the matching.

Due to the row-wise normalization described in Section~\ref{sec:fusion_of_features}, swapping query and gallery sets would likely produce different results. Therefore, we consider both cases by applying row-wise normalization to the original matrix (either pairwise-similarity or pairwise-dissimilarity) and to its transposed version. Then, we apply the greedy algorithm outlined above to both normalized matrices and compute the {\it probability of the matching} (for the best greedy sequential matching) for each matrix. 
As the final identity match, we use the pairings provided by the matrix with the higher probability of matching. We apply this approach to individual features and to all combinations thereof.

\begin{algorithm}
\caption{Greedy Algorithm}\label{alg:greedy}
\begin{algorithmic}
\Ensure $\textbf{S}_{|Q_n|\times|G_n|} \in \mathbb{R}^2$ \\ \Comment{$\textbf{S}$ is the match-probability matrix}
$\textbf{S}_{|Q_n|\times|G_n|} = (s_{ij})_{1\leq i \leq |Q_n|, 1\leq j \leq |G_n|}$ \\ \\
$ 0 \leq (s_{ij})_{1\leq i \leq |Q_n|, 1\leq j \leq |G_n|} \leq 1$
\\
\While{$  \underset{1\leq i \leq |Q_n|, 1\leq j \leq |G_n|}{\max} (s_{ij}) \neq -\infty$}
\\
\State $ k, l \gets \underset{1\leq i \leq |Q_n|, 1\leq j \leq |G_n|}{\arg\max} (s_{ij})$
\\
\State $ k \sim l $   \Comment{Match the $k^{th}$ query and $l^{th}$ gallery identity}
\State $ \underset{1\leq j \leq |G_n|}{s_{kj}} \gets -\infty $
\\
\State $ \underset{1\leq i \leq |Q_n|}{s_{il}} \gets -\infty $
\\

\EndWhile
\end{algorithmic}
\end{algorithm}

\section{EXPERIMENTAL RESULTS - RE-IDENTIFICATION}

Here we focus on the {\it re-identification} problem, independent of the people {\it detection} problem. Therefore, we assume that people detections (i.e., bounding boxes) are provided by an annotated dataset. In practice, application of PRID will also be affected by people mis-detections.


\subsection{Dataset}
\label{ssec:PRID_dataset}
In order to validate the proposed algorithms, we use FRIDA \cite{FRIDA}, a fisheye re-identification dataset captured by three time-synchronized overhead fisheye-lens cameras with fully-overlapping FOVs.
There are 20 different identities, more than 240,000 bounding boxes and the total of 6,106 image triplets captured at different times.
For more details, see \cite{FRIDA}.


Among all our proposed methods only deep-learning features need training (CTL). To train and evaluate CTL, we adopt the 2-fold cross-validation methodology that was used in \cite{FRIDA} in which half the identities are in one fold and the remaining half are in the other fold. Although color-histogram and location features do not require training, in order to ensure fair comparison, we adopt the same testing approach that was used for CTL to evaluate the performance of all our methods. We note that in this testing methodology, for a given pair of time-synchronized video frames, all people from one frame (whose identities belong to a fold) are treated as the query set and those from the other frame (again with identities belonging to the same fold) are treated as the gallery set.

\subsection{Implementation Details}
\label{ssec:PRID_implementation_details}

\begin{table*}[htb]
\caption{Performance comparison of PRID on FRIDA dataset for various combinations of deep-learning (DL), color-histogram (CH) and location-based (LOC) features, for both PPD and CBD distance measures. The highest values of QMS and mAP for each camera pair (e.g., ``C.1+C.2'') and for the cumulative (``Cum.'') metric are shown in boldface.}
\label{tab:PRID_res}
\smallskip
\centering
\begin{tabular}{|c|c|c|c|c|c|c|c|c|c|c|c|} 
\hline
\multicolumn{4}{|c|}{} &
\multicolumn{4}{c|}{QMS [\%]} &
\multicolumn{4}{c|}{mAP [\%]} \\
\hline
DL & CH & LOC/PPD & LOC/CBD & C.1+C.2 & C.1+C.3 & C.2+C.3 & Cum. &
C.1+C.2 & C.1+C.3 & C.2+C.3 & Cum. \\
\hline
\checkmark &  &  & & 80.34 & 91.12 & 66.89 & 79.50 & 83.31 & 90.78 & 76.00 & 83.40\\
\hline
 & \checkmark &  & & 60.41 & 85.84 & 44.91 & 63.80 & 67.68 & 85.60 & 58.46 & 70.63\\
\hline
\checkmark & \checkmark &  & & 83.58 & 94.79 & 70.65 & 83.06 & 84.82 & 93.58 & 76.55 & 85.02\\
\hline
 &  & \checkmark & & 94.76 & 95.83 & 92.51 & 94.37 & 94.84 & 96.78 & 93.97 & 95.20\\
\hline
 & \checkmark & \checkmark & & 95.41 & 96.33 & 93.66 & 95.14 & 95.13 & 97.12 & 94.24 & 95.50\\
\hline
\checkmark &  & \checkmark & & 97.09 & 98.25 & 95.15 & 96.84 & 95.78 & 97.73 & 94.83 & 96.12\\
\hline
\checkmark & \checkmark & \checkmark & & 97.31 & {\bf 98.42} & 95.45 & 97.07 & 95.92 & 97.88 & 94.93 & 96.25\\
\hline
 &  & & \checkmark & 96.63 & 95.01 & 93.28 & 94.98 & 98.18 & 97.92 & 96.92 & 97.68\\
\hline
 & \checkmark & & \checkmark & 97.23 & 96.37 & 94.67 & 96.10 & 98.48 & 98.27 & 97.22 & 98.00\\
\hline
\checkmark &  & & \checkmark & {\bf 98.07} & 97.66 & 96.05 & 97.27 & {\bf 98.83} & 99.24 & {\bf 97.63} & {\bf 98.57}\\
\hline
\checkmark & \checkmark & & \checkmark &  98.03 & 97.75 & {\bf 96.22} & {\bf 97.34} & 98.66 & {\bf 99.26} & 97.18 & 98.37\\
\hline

\end{tabular}
\end{table*}

%
In \cite{FRIDA}, six state-of-the-art deep-learning PRID methods were evaluated on FRIDA. The results show that CTL \cite{CTL} with a ResNet \cite{ResNet} backbone outperforms other methods. This is primarily because other methods have been specifically tailored to traditional rectilinear PRID and do not transfer well to cross-frame fisheye PRID. In contrast, ResNet is designed for generic image inference tasks which makes it adaptable to different applications.
%
%
Thus, we opted for CTL with a ResNet backbone to extract the deep-learning features.

The CTL model has over 27 million parameters. We trained it on NVIDIA Tesla V100 GPUs using Adam optimizer with a learning rate of 3.5e-4, weight decay of 5e-4 and momentum of 0.937 over 300 iterations (typical values used for training deep neural networks). In order to match the bounding box size to the one CTL accepts, we applied zero padding to keep the aspect ratio and then resized the image.

When computing color histograms, we used 256 bins and normalized each histogram to sum up to 1. For location-based features, we used parameters proposed in \cite{josh}: an average person's height of 168cm (in the United States) in the PPD metric and 21 heights ranging from 128cm to 208cm (covering the vast majority of people) in the CBD metric. In softmax normalization of features, we used a temperature value of $T=10$ that was determined through a grid search.

In terms of computational complexity, it takes 30ms on average for CTL inference to extract features of a batch of 32 person-images (256$\times$128 pixels) and 0.54ms to compute a color histogram from each person-image. To map the coordinates of a detected person from one camera to another it takes 12$\mu$s on average. Finally, it takes on average 36$\mu$s to perform a single match by the greedy algorithm. Thus, on average, it takes about 3ms to extract all features of two person-images and verify if they match.

\subsection{Evaluation Metrics}
\label{ssec:PRID_evalutation_metrics}

To measure an algorithm's performance,  we use the {\it Query Matching Score} (QMS) and {\it mean Average Precision} (mAP), similarly to \cite{josh}. QMS is a modified version of the commonly-used {\it Correct Matching Score} (CMS) \cite{CMS} 
that accounts for a potential lack of match, and is defined as follows, 
%
\begin{equation}\label{eqn:qms}
  QMS = \frac{\sum_{n=1}^{N} \sum_{q\in Q_n} \mathbbm{1}(q=\widehat{q})}{\sum_{n=1}^{N} \big|Q_n \cap G_n\big|} 
\end{equation}
where $N$ is the number of time instants (tuples of frames) and $\widehat{q}$ is the predicted identity of query $q$. QMS calculates the ratio of the number of correctly matched query elements to the number of possible correct matches. It is important to note that although cameras have fully-overlapping FOVs, the number of people in a query frame might be different than that in a gallery frame (captured at the same time). This may happen if a full-body occlusion takes place (a person may not be visible in some cameras views) and is reflected in the denominator (\ref{eqn:qms}) which expresses the maximum number of possible correct matches.

We also modified the mAP compared to the one typically used in PRID, where each query is guaranteed at least one match. Since in our PRID at {\it most} one match can occur for each query, we exclude identities that do not have a match in the gallery set.



\subsection{Results}
\label{ssec:PRID_results}

In this section, we report the PRID performance for deep-learning features (DL), color histograms (CH) and location-based features (LOC) separately, and also for three combinations of two feature types (DL+CH, CH+LOC, DL+LOC) and for all three feature types together (DL+CH+LOC) for two distance metrics: PPD 
and CBD.
All results are summarized in Table~\ref{tab:PRID_res}.

We note that since we are using an annotated dataset, each bounding box corresponds to a person truly visible in a camera's FOV.
This allows us to demonstrate the re-identification performance of each algorithm without the confounding influence of errors in people detection.


In addition to reporting camera-pairwise QMS and mAP values (e.g., ``C.1+C.2''), we also report cumulative values (``Cum.'') computed by using the total number of correct matches and the total number of {\it possible} correct matches over all camera pairs rather than for a single pair.

It can be observed, that among individual features the location-based one significantly outperforms DL and CH. Furthermore, the color histogram performs at least 
12\% points below DL
in cumulative QMS and mAP. Clearly, color is hardly enough to distinguish people, but low-level features in DL (that include color) perform much better. However, a person's location captured by time-synchronized cameras is a good indicator of who is who.

Combining the appearance-based features (DL+CH), improves DL's performance by about 4\% points in cumulative QMS and about 2\% points in cumulative mAP. Note, that in FRIDA there are 243,439 bounding boxes. An improvement of cumulative QMS by 4\% points roughly corresponds to an additional 9,700 bounding boxes being correctly re-identified. However, the DL+CH combination still performs well below location-based PRID. Interestingly, for Camera 1-Camera 3 pair the QMS performance of DL+CH combination is close to the one for the location-based algorithms (LOC/PPD and LOC/CBD).
This is so because these two cameras are closer to each other in the physical world compared to the other two camera pairs (see \cite{FRIDA} for camera layout). Thus, the resolution difference for a person positioned between these two cameras is the smallest. This makes appearance-based PRID easier compared to the other camera pairs. Indeed, the DL+CH combination performs worst for Camera 2-Camera 3 pair which are farthest apart. The location-based approach (single feature) also performs better when the cameras are closer to each other, but the performance difference between the best and the worst case is no more than 4\% points in terms of QMS or mAP (compared to over 24\% points QMS and over 17\% points mAP for DL+CH).

The most significant performance boost, when using two feature types, comes from the combination DL+LOC as these are the two best performing approaches individually. Combining LOC/PPD with DL boosts its performance by 2.47\% points in terms of cumulative QMS, which corresponds to about 6,000 additional bounding boxes being correctly re-identified. Similarly, combining LOC/CBD with DL delivers performance boost of 2.29\% points (5,500 additional correctly re-identified bounding boxes) in cumulative QMS. The performance improves further, although very slightly, when color histogram (CH) is combined with DL+LOC features.
%
When we use PPD distance metric, 
performance reaches 97.07\% in cumulative QMS and 96.25\% in cumulative mAP,
whereas 
when we use CBD distance metric
it achieves 97.34\% and 98.37\%, respectively.
However, one should note that the addition of color feature to the DL+LOC/CBD combination 
can have a slightly detrimental effect.

Our results also support conclusions reached in \cite{josh}, namely that algorithms that involve the CBD location metric perform better than the ones that involve the PPD metric. Indeed, LOC/CBD outperforms LOC/PPD by 0.61\% points of cumulative QMS (1,400 additional correctly re-identified bounding boxes). However, when we compare 
DL+CH+LOC/PPD 
and DL+CH+LOC/CBD  
in Table~\ref{tab:PRID_res} 
the performance gap between the two decreases to 0.27\% points of cumulative QMS (650 additional correctly re-identified bounding boxes). Considering that PPD is around 17 times faster than CBD (see \cite{josh} for a computational complexity analysis), 
it seems a better option for real-time system implementation. 

\section{Experimental Results - People Counting}
\label{sec:counting_results}

In Section~\ref{ssec:PRID_results}, we reported PRID performance on FRIDA, where \textit{human-annotated ground-truth} bounding boxes were available for each person in all frames. 
%
%
However, in real-world scenarios ground-truth bounding boxes are not available.
In this section, we focus on the application of PRID to {\it people counting} within a real-life, large-space scenario where multiple fisheye cameras are needed to cover the whole area. Since a person may be visible in several camera views, PRID is essential to avoid over-counting. Moreover,
in this realistic scenario, people-detection algorithms will typically have miss-detections or false-detections. Therefore, people-counting results presented in this section are impacted by both people detection and people re-identification errors.

\subsection{Dataset}
\label{ssec:counting_dataset}

\begin{figure*}[t!]
  \begin{minipage}[c]{0.5\linewidth}
    \centerline{\includegraphics[width=0.76\textwidth]{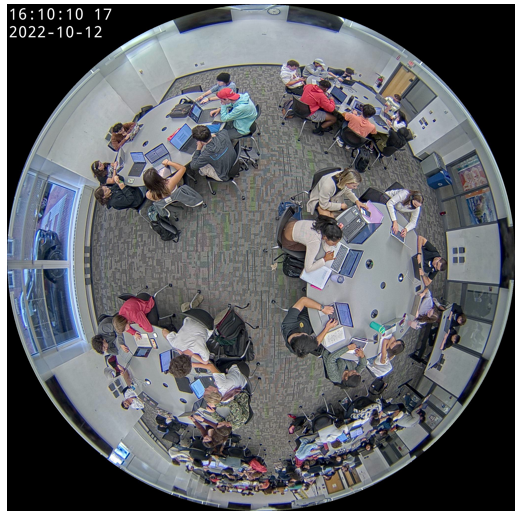}}
    \centerline{(a) Camera A}
  \end{minipage}\hfill
  \begin{minipage}[c]{0.5\linewidth}
    \centerline{\includegraphics[width=0.76\textwidth]{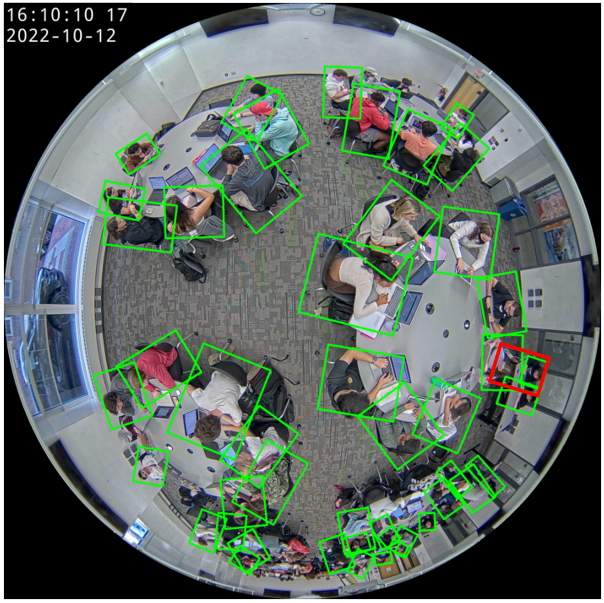}}
    \centerline{(c) Camera A}
  \end{minipage}
  \medskip

  \begin{minipage}[c]{0.5\linewidth}
    \centerline{\includegraphics[width=0.76\textwidth]{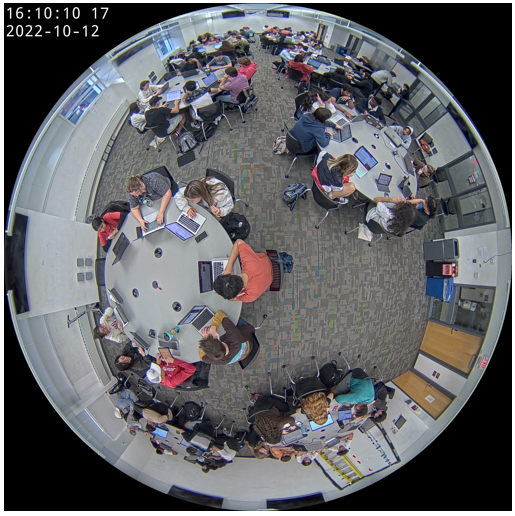}}
    \centerline{(b) Camera B}
  \end{minipage}\hfill
  \begin{minipage}[c]{0.5\linewidth}
    \centerline{\includegraphics[width=0.76\textwidth]{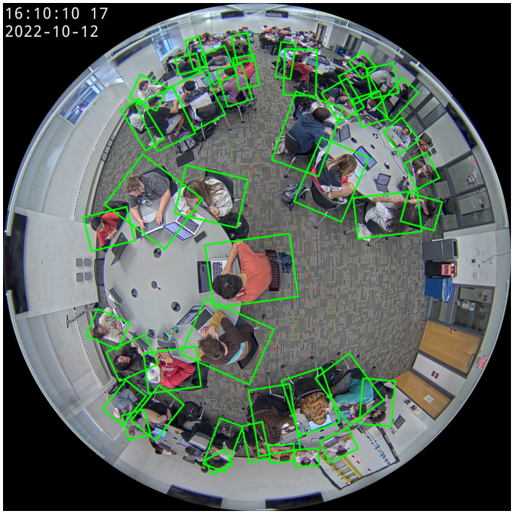}}
    \centerline{(d) Camera B}
  \end{minipage}
  \caption{(a-b) Sample frames from two overhead fisheye cameras overlooking a large classroom; (c-d) The same frames with people detections by RAPiD \cite{RAPiD}.}\label{fig:sample_fisheye}
\end{figure*}

We recorded synchronized videos from 2 overhead fisheye cameras in a large classroom (over 2,000 square feet) spanning 3 lectures, with up to 80 occupants. A sample frame pair from this recording is shown in Figure~\ref{fig:sample_fisheye}(a-b). At each time instant, we counted the number of people present in the classroom by visually inspecting each frame pair. This provides ground-truth people count for performance evaluation.

\subsection{Implementation Details}
\label{ssec:counting_implementation_details}

Person re-identification was implemented using the same setup as reported in Section~\ref{ssec:PRID_implementation_details}.

In order to detect people, we used RAPiD \cite{RAPiD}, a state-of-the-art people-detection algorithm specifically designed for overhead fisheye images. Figure~\ref{fig:sample_fisheye}(c-d) shows people detections produced by RAPiD for the frame pair from Figure~\ref{fig:sample_fisheye}(a-b). One can see that some people are not detected at FOV periphery (e.g., top of the view in Figure~\ref{fig:sample_fisheye}(d)). Occasionally, the algorithm produces false detections (e.g., red bounding box in Figure~\ref{fig:sample_fisheye}(c) contains two people who already have their own green bounding boxes).
Due to these errors and occlusions, the number of bounding boxes detected at the same time in different camera views can be quite different. 
%

To count people, one could compute an average of counts 
from the two cameras, but this could result in undercounting if there are people visible in one camera but not the other (e.g., due to occlusions). On the other hand, adding the two counts could result in overcounting due to the double-counting of people who are visible in both cameras. These effects could be further compounded by people-detection errors. 
A principled approach is to {\it re-identify} people between different camera views and count each person only once. 

Motivated by the above considerations, we apply a slightly modified version of our PRID algorithm (described below) to the bounding boxes detected by the RAPiD algorithm in two camera views and estimate the people count at time $n$ as follows:
\begin{equation}\label{eqn:cn}
  \widehat{C}_n = |\widehat{Q}_n| + |\widehat{G}_n| - \widehat{r}_n
\end{equation}
where $\widehat{Q}_n$ and $\widehat{G}_n$ are the sets of bounding box detections at time $n$ in the query and gallery frames, respectively,  produced by the RAPiD algorithm, and $\widehat{r}_n$ is the number of PRID matches at time $n$. 
Thus, the estimated people count at time $n$ equals the sum of counts obtained from two frames reduced by the number of matched identity pairs in them. 
%


In order to account for people-detection errors, which can result in very different numbers of detected bounding boxes in each camera's view, we slightly modify the matching part of our PRID algorithm. In the matching algorithm discussed in Section\ \ref{ssec:greedy}, the greedy algorithm was exhaustive, meaning it was applied until there are no matches left in the score matrix (i.e., step (3) in Section\ \ref{ssec:greedy}). 
Instead, now we stop the greedy algorithm when there are no more matrix elements greater than some threshold $\tau$ and treat the remaining matrix elements (conditional probabilities) as corresponding to unlikely identity matches.

The threshold $\tau$
controls the trade off between the number of matched and unmatched bounding boxes between the two views. In an ideal scenario, when all occupants in a space are detected in both query- and gallery-camera views, $\tau$ = 0 will force a match of every person in the query set to a person in the gallery set. However, as we have discussed, in practice some occupants may not be detected in one of the views or there may be false detections. 
%
%
In this case, some query or gallery elements may not have a match and $\tau > 0$ is needed to stop the matching process. Thus, $\tau$ serves as a match-probability threshold below which a match is unlikely.

In our people-counting experiments, we treat $\tau$ as a tuning parameter and find its best value for each method (in terms of MAE, as defined in (\ref{eqn:mae})) by searching among a finite set of uniformly-spaced choices over the interval $[0,1]$. We found that the best values for different methods range from 
$2.16\times10^{-4}$ to $0.5$.

\subsection{Evaluation Metric}
\label{ssec:counting_evalutation_metrics}

To measure people-counting performance, we use the {\it Mean-Absolute Error} (MAE) between the true-people counts $C_n$ and their estimates $\widehat{C}_n$ (\ref{eqn:cn}) as follows:
\begin{equation}\label{eqn:mae}
  MAE = \frac{1}{N} \sum_{n=1}^N |C_n-\widehat{C}_n|
\end{equation}


\begin{figure*}[!htb]
  \centering
  \includegraphics[width=0.9\linewidth]{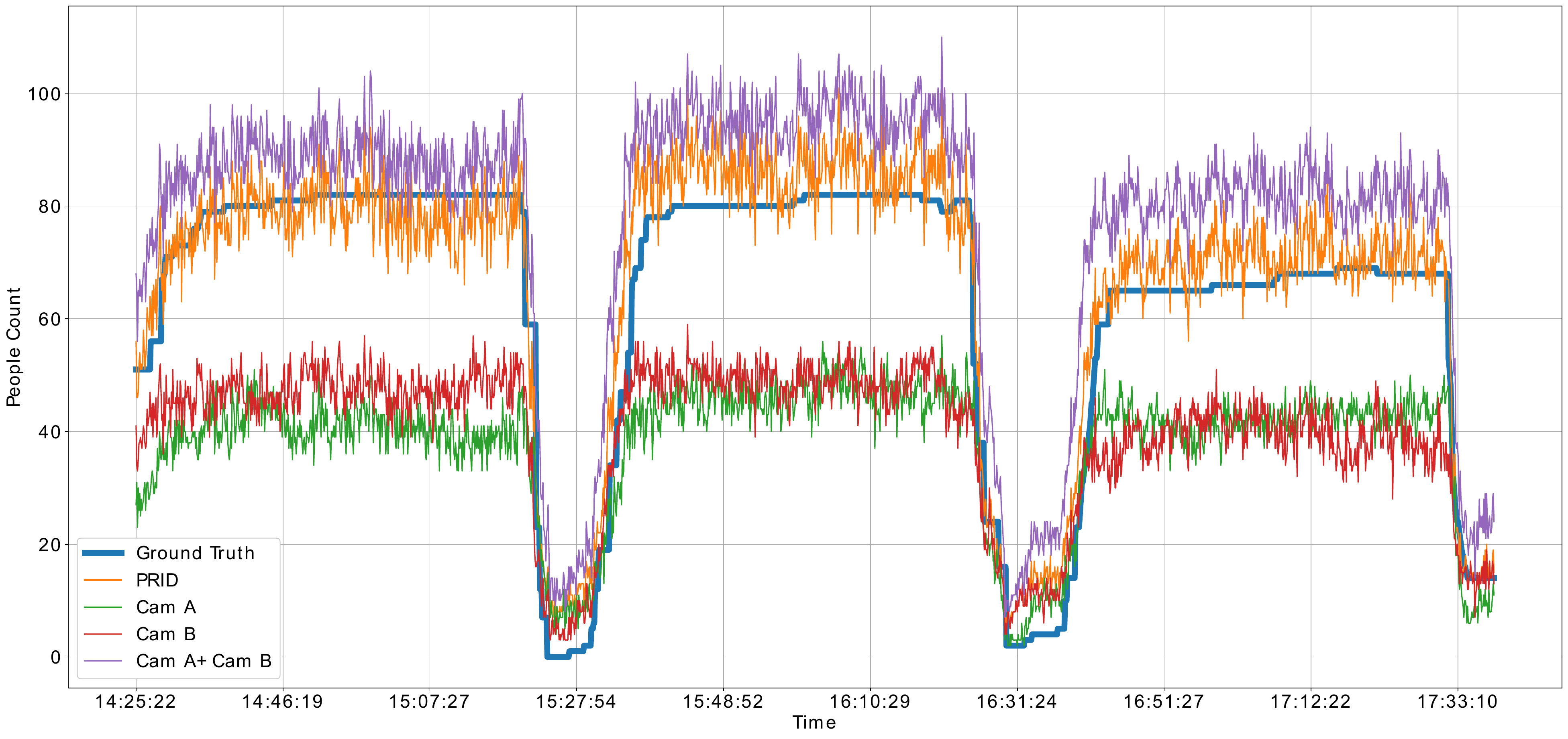}
  \caption{Ground-truth people count and four people-count estimates (two single-camera estimates, sum of single-camera estimates, and an estimate obtained using PRID to eliminate double counts) for a 2-camera recording of 3 lectures (3hr 15min).} 
  \label{fig:people count plot}
\end{figure*}

\subsection{Results}
\label{ssec:counting_results}
 
Figure~\ref{fig:people count plot} shows the time plot of the ground-truth people count and 4 people-count estimates\footnote{The people-count estimates in Figure\ \ref{fig:people count plot} are noisy since they are computed independently for each time instant and no temporal smoothing is applied. While smoothing (e.g., temporal median filtering) could be applied to the estimates, in any real-time application it would have to be causal thus creating a time delay. This smoothness/delay tradeoff has to be carefully adjusted for each practical application.} obtained from our test dataset recorded in a large classroom over a period of 3 lectures (Section~\ref{ssec:counting_dataset}). Three estimates were obtained by counting bounding boxes produced by RAPiD, either from camera A frames, or from camera B frames or by adding the two counts. Clearly, the counts from both camera A and camera B (green and red lines, respectively) severely underestimate the true count (blue line) due to missed-detections at FOV periphery (the room is too large for a single camera). Averaging the two counts would still result in severe undercounting. The sum of the counts from both cameras (purple line) significantly overestimates the true count since many people are counted twice. The fourth estimate (orange line) was obtained by first applying RAPiD to detect people and then performing PRID using our best-performing algorithm (DL+CH+LOC/CBD) to people detections in same-time frames from cameras A and B. Clearly, the PRID-based people-count estimate quite accurately tracks the true people count.
%


While the PRID algorithm combining all three features works very well in people counting, it would be interesting to understand the impact of other feature combinations on people-counting performance. Rather than showing plots like the one in Figure~\ref{fig:people count plot} for different feature combinations, we perform this ablation study by reporting MAE (\ref{eqn:mae}) performance in Table\ \ref{tab:MAE_res}. This table shows the lowest MAE value for each feature combination as determined by a grid search for threshold $\tau$.
To assure independence of the selection of $\tau$ from the test dataset, we performed the grid search on a different video recording (1hr 30min) in the same classroom.

Similarly to re-identification experiments, algorithms involving a location-based feature (either LOC/PPD or LOC/CBD) significantly outperform algorithms based on appearance (color-histogram or deep-learning features or combination thereof), with MAE almost halved. Overall, the lowest MAE value is obtained by the DL+LOC/CBD feature combination (deep-learning features plus location) with $MAE$ = 5.509, which is largely consistent with PRID results (Table\ \ref{tab:PRID_res}).
A close second, with $MAE$ = 5.557, is the three-feature algorithm (DL+CH+LOC/PPD) which is again consistent with PRID results from Table\ \ref{tab:PRID_res} where it ranks among the top 3 feature combinations in terms of QMS. However, the other three-feature algorithm (DL+CH+LOC/CBD) has a slightly higher $MAE$ = 5.891.
%
%
The single-feature location-based algorithms have much higher MAE values: $MAE$ = 6.470 for LOC/PPD and $MAE$ = 6.093 for LOC/CBD suggesting that an addition of CH and/or DL features can boost performance.

\begin{table}[htb]
\caption{MAE value for each feature combination and the corresponding threshold $\tau$ determined through grid search. 
}
\label{tab:MAE_res}
\smallskip
\centering
\begin{tabular}{|c|c|c|c|c|c|c|} 
\hline
DL & CH & LOC & LOC & MAE & Best $\tau$ \\
 & & (PPD) & (CBD) & & \\
\hline
\checkmark &  &  &  & 10.076 & 0.03 \\
\hline
 & \checkmark &  &  & 11.358 & 0.03 \\
\hline
\checkmark & \checkmark &  &  & 11.283 & $9\times10^{-4}$ \\
\hline
 &  & \checkmark &  & 6.470 & 0.5 \\
\hline
 & \checkmark & \checkmark &  & 6.070 & 0.01 \\
\hline
\checkmark &  & \checkmark &  & 6.074 & 0.01 \\
\hline
\checkmark & \checkmark & \checkmark &  & 5.557 & $2.16\times10^{-4}$ \\
\hline
 &  &  & \checkmark & 6.093 & 0.3 \\
\hline
 & \checkmark &  & \checkmark & 5.605 & $6.4\times10^{-3}$ \\
\hline
\checkmark &  &  & \checkmark & 5.509 & $6.4\times10^{-3}$ \\
\hline
\checkmark & \checkmark &  & \checkmark & 5.891 & $2.16\times10^{-4}$ \\
\hline
\end{tabular}
\end{table}

%

While Table~\ref{tab:PRID_res}
reports performance of re-identification only, Table~\ref{tab:MAE_res} reports a combined performance of people detection by RAPiD and of re-identification. If RAPiD introduces people-detection errors (misses or false detections), people
counts can be incorrect even with perfect PRID. However, even if PRID is imperfect, it may still lead to a correct people count, for example if the total number of matches between cameras A and B is correct but some of the matches
are permuted (e.g., person 1 in camera A is matched to person 2 in camera B and person 2 in camera A is matched to person 1 in camera B). PRID affects people counting only if it produces an {\it incorrect number} of matches between two cameras. In conclusion, whereas PRID errors have a full impact on QMS and mAP values in Table~\ref{tab:PRID_res}, they have only a partial impact on the MAE values in Table~\ref{tab:MAE_res}. Still, the ordering of various feature combinations in terms of people-counting MAE is largely consistent with the ordering in terms of re-identification QMS and mAP values.




One final observation is in order. Examining Tables~\ref{tab:PRID_res}
and \ref{tab:MAE_res}, it is clear that algorithms using appearance features (CH or DL or CH+DL) are significantly outperformed by algorithms that combine them with a location-based feature. However, the performance spread between the latter algorithms in both re-identification and people counting is quite small. Since the computational complexity of calculating location-based features is far lower than obtaining a color histogram (CH) or extracting neural features (DL), in applications sensitive to complexity (e.g., real time) a single-feature algorithm using only location may be a good choice (over 94\% in cumulative QMS and over 95\% in cumulative mAP). Furthermore, it was mentioned in Section~\ref{sec:location} that the calculation of LOC/CBD location features is significantly more complex computationally than that of LOC/PPD features. This suggests that in complexity-critical applications single-feature LOC/PPD algorithm would be most appropriate. However, if best performance is required then location features combined with color histograms and/or deep-learning features are a better option.

\section{CONCLUSIONS}

We proposed a multi-feature PRID framework for time-synchronized fisheye cameras with overlapping fields of view. To the best of our knowledge, this is the first work that explores  combining  appearance- and location-based features for PRID. A key technical contribution of our work is a novel probabilistic feature-fusion methodology for identity matching. 

Our experiments show that methods which utilize location information have a high identity-matching accuracy. However, this requires knowledge of camera parameters (both intrinsic and extrinsic). In some scenarios, this information may not be available. Appearance-based methods, on the other hand, do not make use of such information and can be applied to any camera type and any camera-layout topology. However, such methods lag performance-wise behind those that use location-based features. Still, appearance-based features are valuable and do provide a boost to the identity-matching performance (and to a lesser degree people-counting performance) when combined with location-based features. 
%
Clearly, there is still much room for improvement in appearance-based PRID using overhead fisheye cameras. We hope our work will inspire more research in this direction.

\vspace{-1ex}



{\small
\bibliographystyle{ieee}
\bibliography{egbib}
}

\end{document}